# CROSS-DEMOGRAPHIC PORTABILITY OF DEEP NLP-BASED DEPRESSION MODELS

*Tomek Rutowski, Elizabeth Shriberg, Amir Harati, Yang Lu, Ricardo Oliveira, Piotr Chlebek*

Ellipsis Health, Inc., San Francisco, CA, USA

## ABSTRACT

Deep learning models are rapidly gaining interest for real-world applications in behavioral health. An important gap in current literature is how well such models generalize over different populations. We study Natural Language Processing (NLP) based models to explore portability over two different corpora highly mismatched in age. The first and larger corpus contains younger speakers. It is used to train an NLP model to predict depression. When testing on unseen speakers from the same age distribution, this model performs at AUC=0.82. We then test this model on the second corpus, which comprises seniors from a retirement community. Despite the large demographic differences in the two corpora, we saw only modest degradation in performance for the senior-corpus data, achieving AUC=0.76. Interestingly, in the senior population, we find AUC=0.81 for the subset of patients whose health state is consistent over time. Implications for demographic portability of speech-based applications are discussed.

## 1. INTRODUCTION

The incidence of depression is increasing globally, with an estimated 300 million current cases worldwide [1][2][3]. Depression causes a heavy economic and societal burden [4] and is presently exacerbated by COVID-19 [5]. Because depression is under-diagnosed, there is a pressing need for efficient, cost-effective screening and monitoring [6]. Digital health applications can play an important role in remote approaches to screening and monitoring. For fully automated systems, natural-language-based applications are promising. Studies of both spoken [7][8][9] and written [10][11] language show associations between depression and language patterns. Spoken language also provides acoustic and prosodic cues, allowing machine learning models to make sophisticated final predictions [12][13][14][15][16][17][18][19].

Deep learning models are rapidly gaining interest for real-world applications. A gap in current literature, however, is how well such models generalize over different populations. Due to limited publicly-available data sets [20][21], little is understood about the cross-demographic portability of models; most studies test on data from corpora similar to those they are trained on. Collecting and sharing large labeled datasets for depression is a challenge for multiple reasons, including patient privacy. It is thus useful to understand the extent to which models developed on one population will work for another. Because deep learning models require large amounts of data to train, we are interested in portability without model retraining. In this study, we seek to understand depression model portability for basic demographic factors, including age, gender, and ethnicity.

We evaluate the portability of a state-of-the-art NLP-based depression prediction system. The model was trained on a large corpus from a younger population. We test this model on an age-matched dataset as well as data collected from patients in a retirement community. In addition, we examine model generalization to gender and ethnicity subpopulations.

## 2. CORPORA

Our work uses two corpora of American English speech collected by Ellipsis Health. Despite the availability of common corpora used for shared evaluations on depression prediction [22][23], it was necessary to use our own datasets for two important reasons. First, our sets needed to match in terms of how they were collected. Otherwise, demographic differences would be confounded with speech elicitation methodology. Second, we needed a larger number of speakers than were available in the common corpora. In both of our corpora, patients were incentivized financially. The users interacted with an app that posed questions on different topics such as "concerns" and "home life." Users answered by speaking freely. Corpus statistics are given in Table 1.

Our large *General Population (GP)* corpus contains over 15k sessions of speech. An earlier, smaller version of this corpus was used in [24]. Train and test partitions contain no overlapping speakers. Users range in age from 18–65, with a mean age of 30. Spoken responses average about 160 words. Users responded to 4–6 (mean 4.5) questions per session.

**Table 1** Data characteristics. Test data is in italics.

|  | *GP* | | *SP* | |
|---|---|---|---|---|
|  | Sessions | Subjects | Sessions | Subjects |
| *dep+* | 2,563 *653* | 1,836 *653* | *208* | *39* |
| *dep-* | 8,100 *2,425* | 5,483 *2,425* | *479* | *80* |
| *dep +/-* | 2,209 | 526 |  | *42* |
| *Total* | 12,872 *3,078* | 7,845 *3,078* | *687* | *161* |
| RESPONSES: | GP Sessions | | SP Sessions | |
| *Length (words)* | ~800 | | ~450 | |
| *Number/Session* | 4.5 | | 6.1 | |

We refer to the second, smaller collection as the Senior Population (SP) corpus. The SP corpus was collected in Southern California through an Ellipsis partner. The partner is associated with a group practice site comprising over 100 primary care doctors and 200 specialists caring for more than 40,000 patients.

A positive sample of the GP and then SP response is provided below:

> "Right now I'm living with my husband and my two daughters and I am a stay-at-home mom I lost my job when I was pregnant with my second daughter and honestly it's been a roller coaster expected to be fine at home take care kids but it's the most stressful thing I've ever done and I cannot wait to go back to work."

> "My home life is good and it with my sister we have an amazing relationship and we don't get upset with each other very often and when we do get upset with each other were always able to find a medium a meeting area that we can work together to get beyond it or to just realize that it's okay to sometimes be different have a great home life."

A striking difference in demographics between the corpora is age. As shown in Figure 1, age distributions for the two corpora are largely non-overlapping. During response collection, most patients were expected to participate in a session once a week for six weeks. During GP response collection, participation in more than one session was voluntary and could be performed at any interval of at least one week. Repeated speakers in the GP set were always placed only in the training partition.

As stated in Table 1, the SP Dataset contains over 600 sessions. These sessions are shorter than the GP sessions, and contain on average only 450 words versus 800 words in those of the GP. Thus, the mean number of responses is slightly higher (6.1 vs 5.2). Given the size of this corpus, we use it only for testing in this study.

For both corpora, labels came from self-reports using the PHQ-8 (PHQ-9 with the last question removed). The PHQ-8 was collected at the end of each session. We used binary classes for this study. PHQ-8 scores less than 10 were mapped to dep-; scores at or above 10 were mapped to dep+, following [25]. We also introduced a label dep+/- which describes subjects with two or more sessions who had at least one dep+ session and one dep- session. We refer to these patients as "inconsistent" (no judgment intended) with respect to depression class over time; we refer to patients with only same-class sessions over time as "consistent."

### 2.1. Gender Distribution

The two corpora differ naturally in gender distribution. As shown in Table 2, there are more females than males in both cases, but the differences are larger for the SP set.

**Table 2** Gender percentage by corpus

|  | *GP* | *SP* |
|---|---|---|
| *Female* | 58.99% | 62.11% |
| *Male* | 41.00% | 37.88% |

### 2.2. Label Distribution

As described earlier, our data labels represent PHQ-8 questionnaire results. Figure 2 shows the PHQ score label distribution of the GP and SP dataset. They are remarkably similar, with lower rates at very low PHQ-8 scores being allocated somewhat evenly across the rest of the range. The prevalence differences are small (30% in SP; 26.7 in GP).

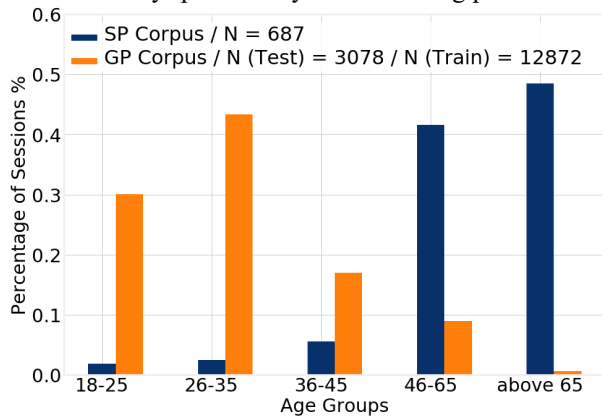

**Figure 1** Normalized age distribution for GP and SP

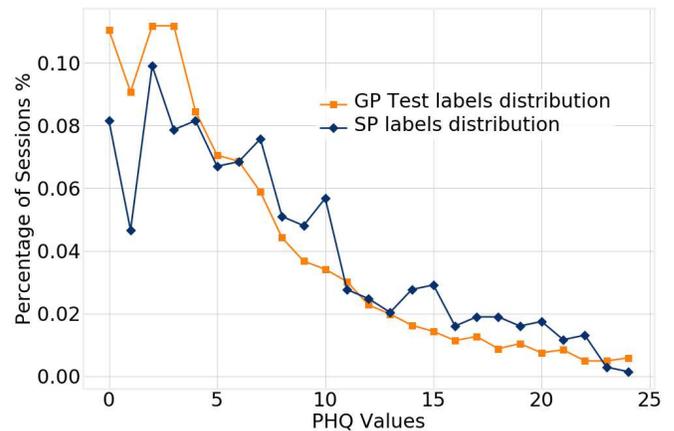

**Figure 2.** PHQ distribution

## 3. DEEP LEARNING MODEL: NLP WITH TRANSFER LEARNING

Our prediction model uses a deep learning language model with transfer learning [28]. Various methods can be used for tokenization [32][33]. We use the spaCy tokenization library; each word is represented by a unique ID. Our corpus word dictionary contains over 20 000 individual tokens. Given the overall shortage of data in this domain, we also tried simpler models including SVM classifiers based on word embeddings [26][27]. With our current training set size, the deep learning approach gave better results. We note however that SVMs gave better results when our training data was roughly 10% of the current size.

We use a language model topology inspired by the AWD-LSTM architecture in [29]. For model fine tuning, we use a method following ULMFiT [30]. Among alternative algorithms, this approach gave strong performance for a single model. As in [29], we use DropConnect for hidden-to-hidden layers; this differs from the dropout approach by deactivating certain weights instead of applying activation mechanisms. We also employed variational dropout, in which the mask is not regenerated when it is called, and embedding dropout, in which occurrences of certain words are removed during the training stage. Back-propagation through time allows handling of longer language dependencies; this has been highly important in our experiments.

Similar to the approach in [30], we apply discriminative fine-tuning so that different layers of the network use a different learning rate. We use a slanted triangular learning rate in which the rate of change is dependent on the stage of the training process. To avoid model forgetting, we use a gradual unfreezing method, partially unfreezing layer by layer.

The core NLP model is trained on multiple data sources including Wikipedia [31]. For the depression prediction task, we retrain the language model on our depression corpora without using labels. Our word transcriptions come from a third-party ASR service. In order to perform the depression classification task, we add additional layers on top of the retrained language model.

## 4. RESULTS AND DISCUSSION

### 4.1. Comparison to past work

Given our use of large but proprietary datasets for both training and evaluation, it is useful to provide evidence of how our approaches perform on shared corpora. Because we did not have access to benchmark corpora, we provide the following indirect comparisons. They are on different datasets but, notably, both use PHQ-8 scores as the gold standard label. We chose AVEC 2019 [22] as our comparison to report results here for our NLP-based system only; this is because that is the system described elsewhere in this paper. Since [22] does not provide classification results, we computed regression performance for our models and reported a RMSE metric used in [22] and elsewhere for the same data set. RMSE is an error metric and inversely correlated with performance.

**Table 3** Indirect performance comparison. AVEC, GP, and SP are different data sets using PHQ-8 labels.

|  | Regression RMSE / MAE | Classification ROC AUC |
|---|---|---|
| *AVEC Test* | 5.510 / 4.200 | |
| *GP Test* | 4.241 / 3.176 | 0.828 |
| *SP* | | 0.761 |

As shown, the results for our NLP system demonstrate lower RMSE than the results for the system in [34], again on different data. For comparison to the following sections, Table 3 also provides our standard binary classification results for both GP and SP test sets. We believe that ROC AUC [35] is the best metric for our use case where the data is imbalanced and class separation is relevant.

### 4.2. Model portability

We now turn to classification results on our data. Figure 3 shows binary classification results for different test sets; the NLP system is always trained on the GP corpus training data.

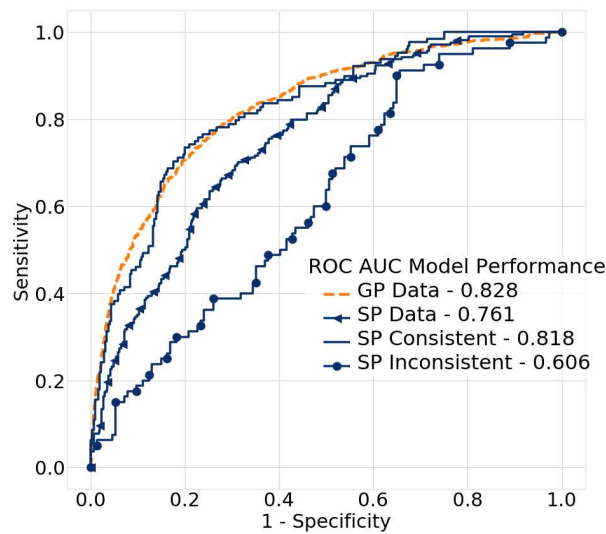

**Figure 3** ROC AUC performance on various test sets. The model is always trained on GP training data.

GP indicates model performance when the GP-trained model is tested on the separate GP test set. SP indicates performance when the GP model is tested on SP data without retraining or tuning. As shown, there is a performance degradation (from 0.828 to 0.761) associated with the mismatch. Given the major age distribution difference, this is a better result than we expected. It suggests that patterns captured by the language model as indicators of depression may be largely portable across these age groups without need for significant retraining.

### 4.3. Performance by patient class consistency

In the SP test corpus, patients participated in a longitudinal study as described earlier. We discovered, interestingly, that classification performance of the GP-trained NLP model depends strongly on consistency of a patient's self-reported PHQ-8 scores over the multiple-session collection (with sessions roughly one week apart on average). Out of 161 unique patients in the SP corpus, 119 had PHQ-8 scores that were either always dep- or always dep+. The remaining 42 patients had a mix (dep+/-) of sessions over the course of data collection. The mean of responses per session over the full SP set is 6.1, as shown in Table 1; for the longitudinal-study patients, this decreased to 4.2. Interestingly, we found a large difference in the number of responses per session between consistent and inconsistent longitudinal-study SP users. Consistent users averaged 3.8 responses, whereas inconsistent users averaged 5.5. This difference is not correlated with depression class. This difference is relevant to the design of future speech elicitation applications.

Overall, consistent patients were more concise and had fewer responses than inconsistent patients. Figure 3 reveals that there is a marked difference in model performance as a function of user consistency even though each session is treated independently. Performance for consistent SP patients is 0.82, versus 0.61 for inconsistent SP patients. Despite the large mismatch in age as well as other factors in the two corpora, the model surprisingly performs about as well on mismatched-age (SP) consistent users as on matched-age (GP) users. We plan to investigate the role of longitudinal consistency in future work. What is clear is that there is good portability in the NLP model, especially for consistent patients.

### 4.4. Performance by age

Table 4 shows the model performance split by age groups. Note that the number of users under 50 in the SP corpus is by design very small. We include performance for these younger speakers regardless, as the training set for these classes is large.

**Table 4** Model performance by age

|  | *GP size* | *SP size* | *AUC GP* | *AUC SP* |
|---|---|---|---|---|
| *18-25* | 853 | 12 | 0.829 | 1 |
| *26-35* | 1393 | 17 | 0.825 | 0.984 |
| *36-45* | 514 | 41 | 0.820 | 0.796 |
| *46-65* | 288 | 289 | 0.813 | 0.782 |
| *above 65* | 23 | 328 | 0.733 | 0.688 |

As shown, performance on the GP Test set is heavily correlated to that of the GP Train set age distribution. The same pattern is obtained for the SP set, although the very low data counts limit interpretability. Thus, we draw conclusions from only the results from the two largest groups (age 46 and above).

For the SP set, we further looked at performance by actual age; this was facilitated by having actual age for this corpus. For each age threshold, we created a group older and a group younger than that threshold, as shown in Figure 4. Because of the distribution of data given in Figure 1, the light line (count of sessions for speakers beyond age 30) decreases in value as age increases. Similarly, the dark line (count for speakers below the age threshold) increases as it collects additional speakers.

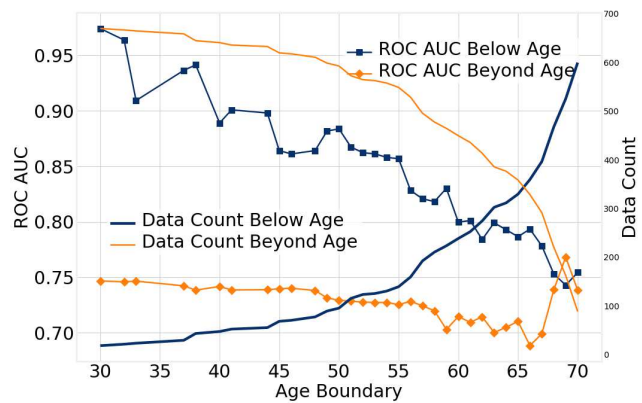

**Figure 4** Age threshold Analysis for SP

Model performance degrades as the age threshold increases (dark squares). This means that it becomes increasingly difficult for the system to perform accurate prediction as the age of the group increases. The sharp increase just before age 70 of the "Beyond Age" trend is noise due to data sparsity. Thus, while we obtain good portability over age (as seen previously), generalization does decline with age. This may be attributable to model fit or to ASR performance or other factors, all areas to explore in future work.

### 4.5. Performance by gender

In addition to age, we also examined gender. In this case, the GP model is tested on gender sub-splits of the GP and

SP sets respectively. As shown in Table 5, performance is remarkably close for females and males for the GP data. For the SP dataset, we see an improvement for females over males.

Table 5 SP and GP performance results by gender

|  | GP | | | | SP | | | |
|---|---|---|---|---|---|---|---|---|
|  | SIZE | AUC | SPEC | SENS | SIZE | AUC | SPEC | SENS |
| FEMALE | 1,799 | 0.827 | 0.760 | 0.760 | 433 | 0.784 | 0.711 | 0.710 |
| MALE | 1,238 | 0.825 | 0.744 | 0.738 | 254 | 0.752 | 0.691 | 0.695 |

### 4.6. Performance by ethnicity

In addition to our focus on age, for the GP Test we also looked at ethnicity groups available in our metadata set. We collected ethnicity for only our GP data. Ethnicity was self-reported. It is likely that self-reported ethnicity categories do not reflect rich information about the level of influence of ethnicity for the individuals in our study. Nevertheless, we did expect to see some variation based on this factor. Results for the GP-trained NLP model were evaluated on ethnicity subgroups in the GP test data. Information is shown in Table 6. Specificity and sensitivity are given at the point of EER; they deviate from each other only due to data sparsity. As with the earlier evaluations, we did not tune or retrain the GP model; it contained a roughly similar ethnicity distribution to the test data shown in Table 6.

Table 6 Results by ethnicity in GP test set

| ETHNICITY | SUBGROUP SIZE | ROC AUC | SPECIFICITY EER | SENSITIVITY EER |
|---|---|---|---|---|
| CAUCASIAN | 2047 | 0.829 | 0.752 | 0.757 |
| HISPANIC | 246 | 0.788 | 0.737 | 0.730 |
| AFRICAN-AMERICAN | 244 | 0.815 | 0.625 | 0.707 |
| MIXED | 170 | 0.856 | 0.778 | 0.770 |
| EAST ASIAN | 125 | 0.819 | 0.752 | 0.750 |
| OTHER | 89 | 0.816 | 0.718 | 0.666 |
| SOUTH ASIAN | 58 | 0.892 | 0.680 | 0.750 |
| CARIBBEAN | 37 | 0.800 | 0.533 | 0.714 |
| DECLINE | 25 | 0.847 | 0.695 | 0.500 |

It is also worth noting that the Hispanic group has lower performance than the other ethnicity groups, a result requiring further analysis. Overall, however, performance for all ethnicity groups remains close to the general model performance, despite lack of any tuning or retraining.

### 5. CONCLUSIONS AND FUTURE WORK

We found that a state-of-the-art depression classifier based on deep NLP and transfer learning showed excellent portability over age, gender, and ethnicity. Using two corpora almost non-overlapping in age but similar in collection design, we found only a small degradation in binary classification performance (0.06 absolute AUC) when testing speakers mismatched for age. Interestingly, this degradation nearly disappeared for those seniors reporting consistent class labels over their longitudinal samples. Despite the overall robustness over age, we do see degradation with increased test sample age for a model trained on younger speakers. We found similar promising generalization when testing a general model on gender and ethnicity subsets.

In future work, it will be important to explore multi-dimensional contextual factors associated with demographic labels. For example, portability will be affected by factors such as ASR performance itself, cultural differences in stigma, differences in illness levels, devices, comorbidity conditions, and many other areas. By better understanding these factors and how they affect model portability, researchers can better design robust systems for behavioral health screening and monitoring in digital health applications.

### 6. ACKNOWLEDGMENTS

We thank David Lin, Mike Aratow, Tahmida Nazreen, Chloe Owen, and Mainul Mondal for support and contributions.